\documentclass[a4paper, 10 pt, conference]{ieeeconf}

\IEEEoverridecommandlockouts                              

\usepackage{csquotes}
\usepackage{times}

\usepackage{graphicx}

\usepackage{url}

\usepackage[maxbibnames=10,sorting=none]{biblatex}
\AtBeginBibliography{\small}

\usepackage{balance} 

\usepackage{multicol}
\usepackage{hyperref}
\usepackage{algorithm,algpseudocode}

\usepackage{capt-of,etoolbox}
\usepackage{blindtext}

\usepackage{graphics} 
\usepackage{colortbl}
\usepackage{color}
\usepackage{mathtools}
\usepackage{amsmath,amssymb,amsfonts}
\usepackage[dvipsnames]{xcolor}
\usepackage{tabularx}
\usepackage{multirow}
\usepackage{colortbl}
\usepackage{color}
\usepackage[belowskip=-15pt,aboveskip=0pt]{caption}

\newcommand{\loosepar}{\looseness=-1}

\title{\LARGE \bf
Fast LiDAR Informed Visual Search in Unseen Indoor Environments} 

\author{Ryan Gupta$^{1}$, Kyle Morgenstein$^{1}$, Steven Ortega$^{2}$ and Luis Sentis$^{1}$
\thanks{$^{1}$Department of Aerospace Engineering and Engineering Mechanics,
        University of Texas at Austin,
        Austin, TX 78712 USA
        {\tt\small ryan.gupta@utexas.edu}}%
\thanks{$^{2}$Department of Mechanical Engineering,
        University of Texas at Austin,
        Austin, TX 78712 USA
}}

\begin{document}

\let\oldtwocolumn\twocolumn
\renewcommand\twocolumn[1][]{%
    \oldtwocolumn[{#1}{
    \begin{center}
    \vspace{-2em}
           \includegraphics[width=\textwidth]{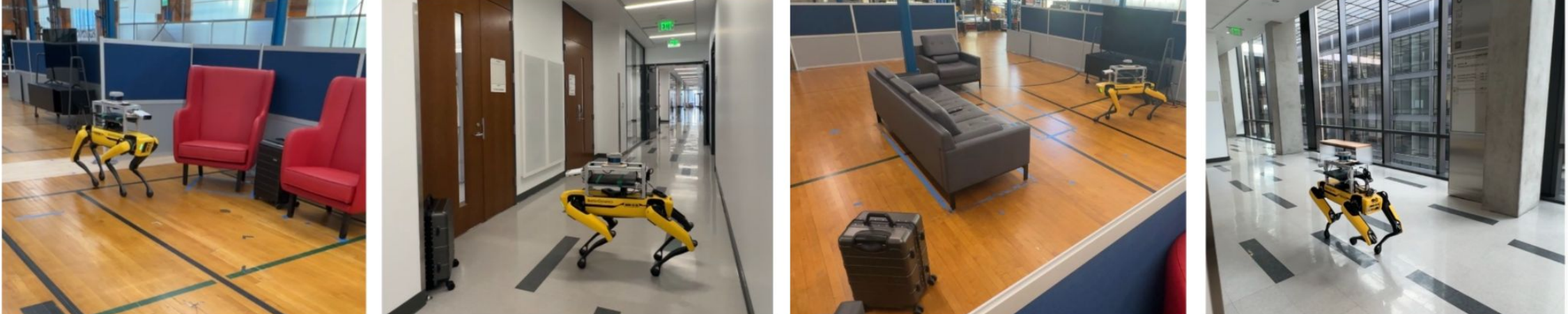}
           \captionsetup{belowskip=1pt}
           \captionof{figure}{Spot quickly finds objects in varied indoor environments. Video of experiments can be found on our website.}
           \label{fig:spot}
           \captionsetup{belowskip=10pt}
        \end{center}
    }]
}
\smallskip

\maketitle
\thispagestyle{empty}
\pagestyle{empty}

\begin{abstract}

This paper details a system for fast visual exploration and search without prior map information. We leverage frontier based planning with both LiDAR and visual sensing and augment it with a perception module that contextually labels points in the surroundings from wide Field of View 2D LiDAR scans. The goal of the perception module is to recognize surrounding points more likely to be the search target in order to provide an informed prior on which to plan next best viewpoints. The robust map-free scan classifier used to label pixels in the robot's surroundings is trained from expert data collected using a simple cart platform equipped with a map-based classifier. We propose a novel utility function that accounts for the contextual data found from the classifier. The resulting viewpoints encourage the robot to explore points unlikely to be permanent in the environment, leading the robot to locate objects of interest faster than several existing baseline algorithms. Our proposed system is further validated in real-world search experiments for single and multiple search objects with a Spot robot in two unseen environments. Videos of experiments, implementation details and open source code can be found at \url{https://sites.google.com/view/lives-2024/home}.\loosepar{}

\end{abstract}

\section{INTRODUCTION}

Planning and real-world execution for autonomous visual exploration are receiving significant attention from the robotics community due to their relevance in a wide range of scenarios including inspection \cite{nguyen2024grey}, search \cite{chen2023hybrid}, disaster response \cite{bi2023cure}, agricultural monitoring \cite{christiansen2017designing}, reconnaissance and surveillance \cite{liang2022reconnaissance}. 
As mobile robots begin to proliferate in our communities, visual planning capabilities remain fundamental for autonomous agents to achieve success in human environments \cite{zheng2023system}.
Active planners in unknown spaces are required to make decisions based on incomplete and noisy information about partial environments.
This requires planners equipped to generate high quality plans under uncertainty.\loosepar{}

Today's robots are frequently equipped with LiDAR sensors that cast a full view of the surroundings.
State of the art visual exploration planners like \cite{best2022resilient,vidal2020multisensor} demonstrate efficiency by fusing LiDAR and visual information.
At the same time, the potential to detect additional information in LiDAR scans for online control has grown with advances in parallel computing.
While methods exist for 3D LiDAR segmentation \cite{qi2017pointnet,zamorski2020adversarial}, they are limited by high computational requirements \cite{jhaldiyal2023semantic}.
In contrast, projection-based methods run online but are used to detect moving obstacles for autonomous vehicles \cite{chen2022automatic,mersch2022receding}. 
The literature also demonstrates planning under reduced uncertainty in unseen environments by exploiting features common to the places they are designed for \cite{saroya2020online,ramakrishnan2020occupancy}.
Despite improvements in efficiency of planning algorithms for both LiDAR and vision sensors \cite{best2022resilient}, it remains an open problem to enhance performance in autonomous planning and execution \cite{wang2019autonomous}. 
To this end, we propose to exploit features common to many indoor environments using a lightweight scan segmentation method for efficient visual search planning.\loosepar{}


The key insight in this work is to exploit contextual information available in wide Field of View LiDAR scans to filter out points unlikely to be the search target.
This deviates from traditional exploration methods, which focus on completeness.
Instead, the proposed method prioritizes regions more likely to be the search target in the robot's surroundings.
We hypothesize exploring agents can improve time efficiency in visual search by exploiting this additional information.
A map based classifier is used to train a model to recognize features inherent to many indoor environments.
The perception model feeds the planner the non-permanent points, \textit{non-map points}.
These points guide the Next Best View (NBV) planner towards possible search target locations, assumed to be in the set of non-map points in the environment.
Examples of search environments are shown in Fig. \ref{fig:spot}.\loosepar{}

The contributions of this work can be summarized as follows:
\begin{itemize}
    \item Achieve significantly faster search indoors than baselines by augmenting multisensor exploration with contextual information from LiDAR scans 
    \item A scan segmentation policy that identifies non-permanent points in unseen, real-world environments
    \item A novel frontier-based exploration formulation that accounts for segmented scan information 
\end{itemize}

\section{Related Works}
\label{sec:related}

The two common approaches to exploration include frontier \cite{yamauchi1997frontier} and information-theoretic sampling methods \cite{charrow2015information}.
Sampling approaches avoid the expensive frontier generation process but require evaluation of every sample, motivating hybrid approaches \cite{dai2020fast}.
Alternatively, \cite{umari2017autonomous} use RRTs to search for frontier points; achieving efficiency using multiple trees and decoupling tree expansion from movement.
Another approach expands RRTs with accessible space and executes the edge with the highest information gain in a receding horizon fashion \cite{bircher2016receding}.
The key difference between these methods and ours is the addition of segmented scans, which is shown to significantly improve task completion time.\loosepar{}

State of the art multisensor exploration is shown with an underwater vehicle \cite{vidal2020multisensor} and multiple UAVs \cite{best2022resilient}.
They combine frontier generation for two sensors and use a single utility function to score frontier candidates for surface inspection. 
We modify their method to sample candidates \textit{at} frontier centroids instead of \textit{surrounding} them.
This modified method is equivalent to our proposed planner \textit{without} the use of segmented scan information and represents a third baseline method.
This baseline clearly demonstrates the positive effect of the proposed contextual information.\loosepar{}

Partially observable markov decision processes (POMDP) are commonly used in visual object search \cite{kaelbling1998planning,zheng2023system}.
The latter presents a platform-generalizable 3D viewpoint planner, capable of reasoning about viewpoints underneath furniture or between appliances.
Their work includes a 2D planner for multi-room search, however, it is not tailored to search larger spaces for things that may be in plain sight; a skill that offers real-world application in search and rescue.
Because our goals involve searching larger spaces quickly, the aforementioned exploration algorithms provide more fair baseline comparisons for the proposed system.\loosepar{}


World prediction is used to learn a model of relevant environments and exploit their structure to explore high-value regions first.
Ref. \cite{saroya2020online} use topological features like loops or dead-ends from a database of subterranean tunnels to inform a frontier based exploration policy. 
Instead, \cite{ramakrishnan2020occupancy}  considers occupancy anticipation from egocentric RGB-D for navigation by an exploration planner. 
Ref. \cite{caley2019deep} learns to predict locations of exits in building with a convolutional neural network using a database of building blueprints.
Our proposed system leverages a similar principle to exploit the structures inherent to many indoor environments - like planar walls, hallways, rooms and loops - in order to guide the planner.
We achieve this in a novel way by segmenting scans to identify non-permanent features in the environment.
Ultimately, the addition of these features are shown to enable our method to find objects of interest more quickly than the examined baselines.\loosepar{}



\section{Methods}
\label{sec:method}

The overall scan classifier and planning module are described in Alg. \ref{algorithm} with key components detailed in this section.\loosepar

\vspace{-0.5em}
\begin{algorithm}[H]
 \caption{Map-Free LiDAR Informed Search()}  \label{algorithm}
 \renewcommand{\algorithmicrequire}{\textbf{Input:}}
 \renewcommand{\algorithmicensure}{\textbf{Output:}}
  \begin{algorithmic}
  \Require {$s_i, x_i$ \\ LiDAR Scan, Robot Pose}
  \Ensure {$x^*_{\text{next}}$ (Next viewpoint)}
   \State $s_i^{\text{class}} = \text{NeuralNetworkClassifier}(s_i)$  
   \State $\mathcal{M}^{\text{LiDAR}}_{i} \gets \text{LiDARMapUpdate}(s_i, x_i, \mathcal{M}^{\text{LiDAR}}_{i-1})$
   \State $\mathcal{M}^{\text{Visual}}_{i} \gets \text{VisualMapUpdate}(s_i, x_i, \mathcal{M}^{\text{Visual}}_{i-1})$
   \State $\{\mathbb{C}^{\text{LiDAR}}\} = \text{GetLiDARFrontiers}(\mathcal{M}^{\text{LiDAR}}_{i})$
   \State $\{\mathbb{C}^{\text{Visual}}\} = \text{GetVisualFrontiers}(\mathcal{M}^{\text{Visual}}_{i})$
    \For{$n \gets 1$ to $N$} \Comment  for each frontier $c_n$ in $\mathbb{C}$
    \For{$j \gets 1$ to $4$} \Comment  for each  viewpoint at $c_n$
   \State $\text{util}_{n,j} = \text{ComputeSampleUtility}(c_n, x_i, s_i^{\text{class}})$
   \EndFor
   \EndFor 
   \State $x^{*}_{\text{next}}$ $\gets$ $\arg \max_{n,j} (\text{util}_{n,j})$ \\
  \Return $x^{*}_{\text{next}}$
 \end{algorithmic}
 \end{algorithm}
 \vspace{-2em}
\subsection{Environment}
The environment is a discrete set of points $\mathcal{E}$, split into two subsets $\mathcal{E}^{\text{free}} \subset \mathcal{E}$ and $\mathcal{E}^{\text{occ}} \subset \mathcal{E}$, representing free and occupied cells.
Both sets are initially unknown to the robot, but are assumed to follow a structure inherent to indoor environments (e.g. hallways, planar walls, open rooms and loops).\loosepar{}

\subsection{Ground Truth LiDAR Scan Classification}
\label{sec:ground-truth}

Ground truth classification uses map $\mathcal{M}$ to divide $\mathcal{E}^{\text{occ}}$ into $\mathcal{E}^{\text{non-map}}$ and $\mathcal{E}^{\text{map}}$.
In this work, $\mathcal{E}^{\text{map}}$ are points in the environment that are deemed to be Long-Term Features (LTFs) during classification, while $\mathcal{E}^{\text{non-map}}$ are points that are either Short-Term Features (STFs) or Dynamic Features (DFs) \cite{biswas2017episodic}.
LTFs represent permanent features while STFs and DFs are non-permanent static and dynamic points, respectively.
Given $\mathcal{M}$, represented as a set of lines $\{l_i\}_{1:n}$, points are classified into these categories as follows.\loosepar{}

Let $x_i$ denote robot pose, and $s_i$ denote observation at time step $t_i$. 
Observation $s_i$ consists of $n_i$ 2D points, $s_i=\{p^j_i\}_{j=1:n_i}$. 
Observations are transformed from local to global frame using affine transformation $T_i\in SE(3)$.\loosepar{}

\subsubsection{LTF}
\label{sec:LTF}
First, an analytic ray cast is performed \cite{biswas2012depth} to determine expected laserscan based on map $\mathcal{M}$ and current robot position $x_i$.
Given observations, the probability that points correspond to the static map can be written:\loosepar{}
 \begin{equation}
 \label{eq:ltf}
 P(p_i^j|x_i,M) = \text{exp}\left(-\frac{\text{dist}(T_ip_i^j,l_j)^2}{\Sigma_s}\right)
  \end{equation}
where $\Sigma_s$ is the scalar variance of observations from sensor accuracy. If Eq. \ref{eq:ltf} is above a threshold, point $p_i^j$ is an LTF.\loosepar{}
  
\subsubsection{STF}
Remaining points are either STFs or DFs. 
Observations at current time $i$, $p_i^j$, are compared with prior observations at time $k$, $p_k^l$ to determine correspondence between points in subsequent observations. 
The likelihood of a point corresponding between time steps is computed as:\loosepar{} 
 \begin{equation}
\label{stf_eqn}
 P(p_i^j,p^l_k|x_i,x_k) = \text{exp}\left(-\frac{||T_ip_i^j-T_kp^l_k||^2}{\Sigma_s}\right)
  \end{equation}
 where $p^l_k$ is the nearest point from $p_i^j$ among points which does not belong to LTF at other timesteps, defined as
 \begin{equation}
 p^l_k = \arg \text{min} ||T_ip_i^j-T_kp^l_k||
 \end{equation}
When Eq. \ref{stf_eqn} is greater than some threshold, point $p_i^j$ is classified as an STF.
Remaining points in $p_i^j$ are classified as DFs.
The result is classified scan $s^{\text{class}}_i$.
Open source code is published by Biswas et. al. \cite{biswas2017episodic} at \cite{enml}.\loosepar{}

\subsection{Map-free LiDAR Scan Classification}
\label{sec:neural}

The classification method defined in \ref{sec:ground-truth} is limited by the necessity of static map information $\mathcal{M}$.
Therefore, this work uses supervised learning to train a model to reproduce binary classifications for $\{p^j_i\}_{j=1:n_i}$ in scan $s_i$, where $\{p^j_i\}_{j=1:n_i,class} \in \{-1, 1\}$

\subsubsection{Dataset and Data Acquisition}
\label{ssec:dataset}

\begin{figure}
    \centering
    \includegraphics[width=\columnwidth]{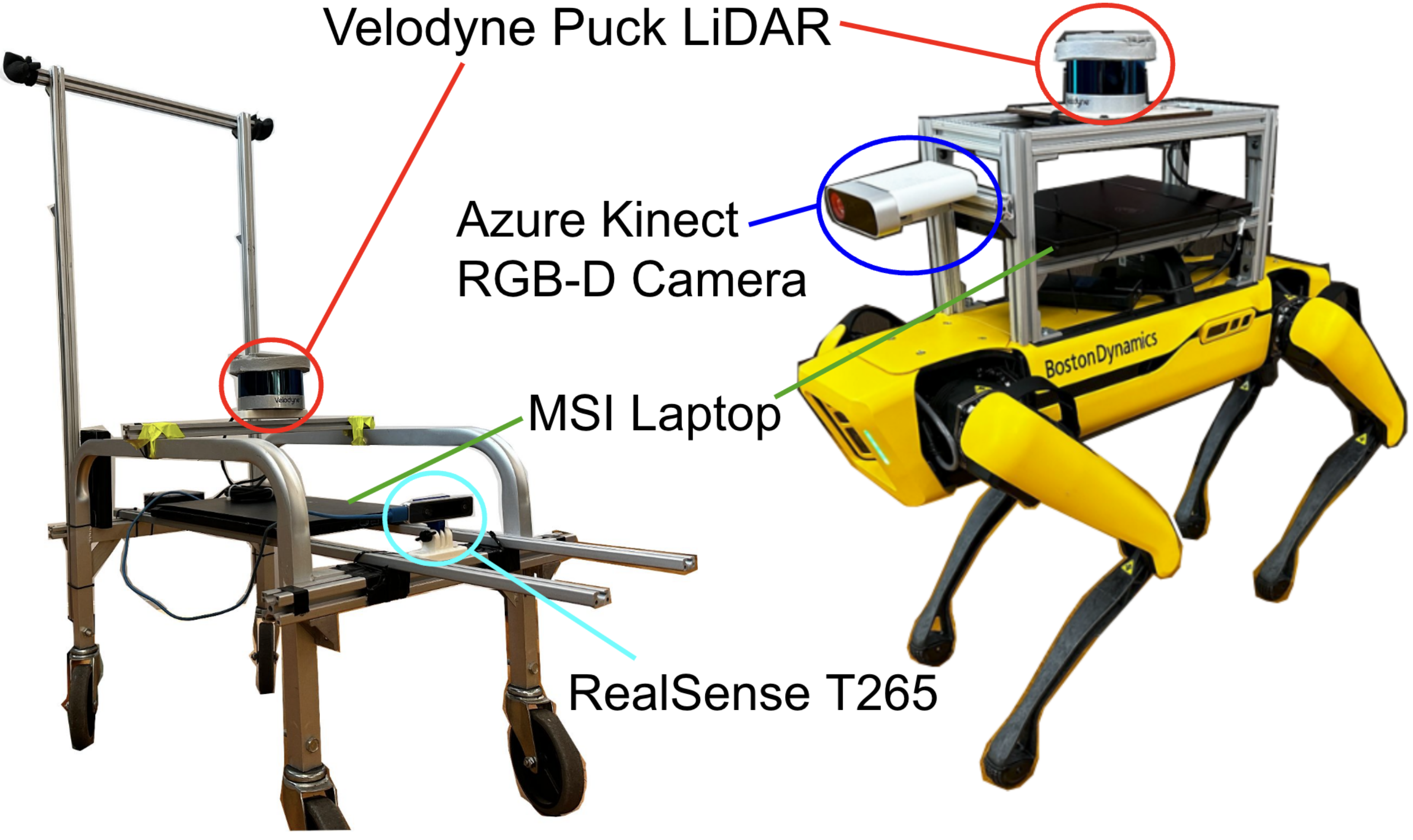}
    \caption{The cart used for labeled data acquisition and the Spot robot used for deployment. The RealSense provides odometry estimate to the cart, required for ground-truth estimation. Spot is equipped with an RGB-D Azure Kinect for detection.\loosepar{}}
    \label{fig:cart}
\end{figure}

We first collect a dataset, however data acquisition with a robot is difficult to scale due to its mechanical limits.
To ease data acquisition we design an easily maneuverable cart platform (see fig. \ref{fig:cart}) to mimic the robot.
Data is labeled and recorded online as the cart is pushed through eight indoor environments on the university campus that vary in size and shape in attempt to capture generic features found in different indoor environments.
Training and test maps can be found on the website.
A laptop with static map information on board the cart receives odometry and scan data from sensors, then uses the method described in the previous section to maintain a pose estimate and collect dataset $\mathcal{D}$, consisting of tuples of data.
The tuples are given as $\mathcal{D} = \{(x_i, s_i,s_i^{\text{class}})\}^{N}_{i=1}$, where $x_i$ represents robot pose, $s_i$ is the raw LiDAR scan data, and $s_i^{\text{class}}$ is the classified scan.
Note that non-map elements in $s_i^{\text{class}}$ are value $-1$ and map elements are value $1$.
$N$ is the total number of data points collected.\loosepar{}

The Velodyne Puck has $n_{i} = 897$ and classified scans are recorded at 5Hz.
A dataset of size $N \simeq 145,000$ is collected in this study, representing roughly 8 hours of time.
Training data can be found on the project website.\loosepar{}

\subsubsection{Architecture and Training}

In practice, the model only receives raw scan data, pose estimates and previously \textit{predicted} labels from the robot during operation and must classify pixel-wise the scan online.
As a result, we formulate pixel-wise classification as a supervised learning problem during training and an auto-regressive problem during inference. 
Given dataset $\mathcal{D}$, we shuffle and batch the dataset between time steps and locations to minimize location-based and temporal bias during training. 
Because no map information is provided explicitly, we provide the model with a history buffer of $k$ time steps for scan ranges and robot poses, and $k-1$ previously estimated ground truth labels. We corrupt the ground truth labels during training by randomly bit-flipping the classification of 10\% of the labels, chosen by sampling indices from a uniform random distribution. 
In order to produce a $k$ length estimated label history input, we take the exponential weighted average of the $k-1$ corrupted pixel-wise classifications to estimate the current timestep's classification.\loosepar{}

During inference, the label history buffer is populated by previous estimates from the policy, cropped by a length $k-1$ sliding-window. 
The history buffer is concatenated with the updated pixel-wise exponential weighted average in the same way as was done during training to obtain the $k$ length estimated label history input. 
At initialization, the policy is bootstrapped with zero-value poses, scans, and labels, and run for $k$ steps of inference until the history buffer is full.
Code for training a model can be found on the project website.\loosepar{}

\begin{figure}
    \centering
    \includegraphics[width=\columnwidth]{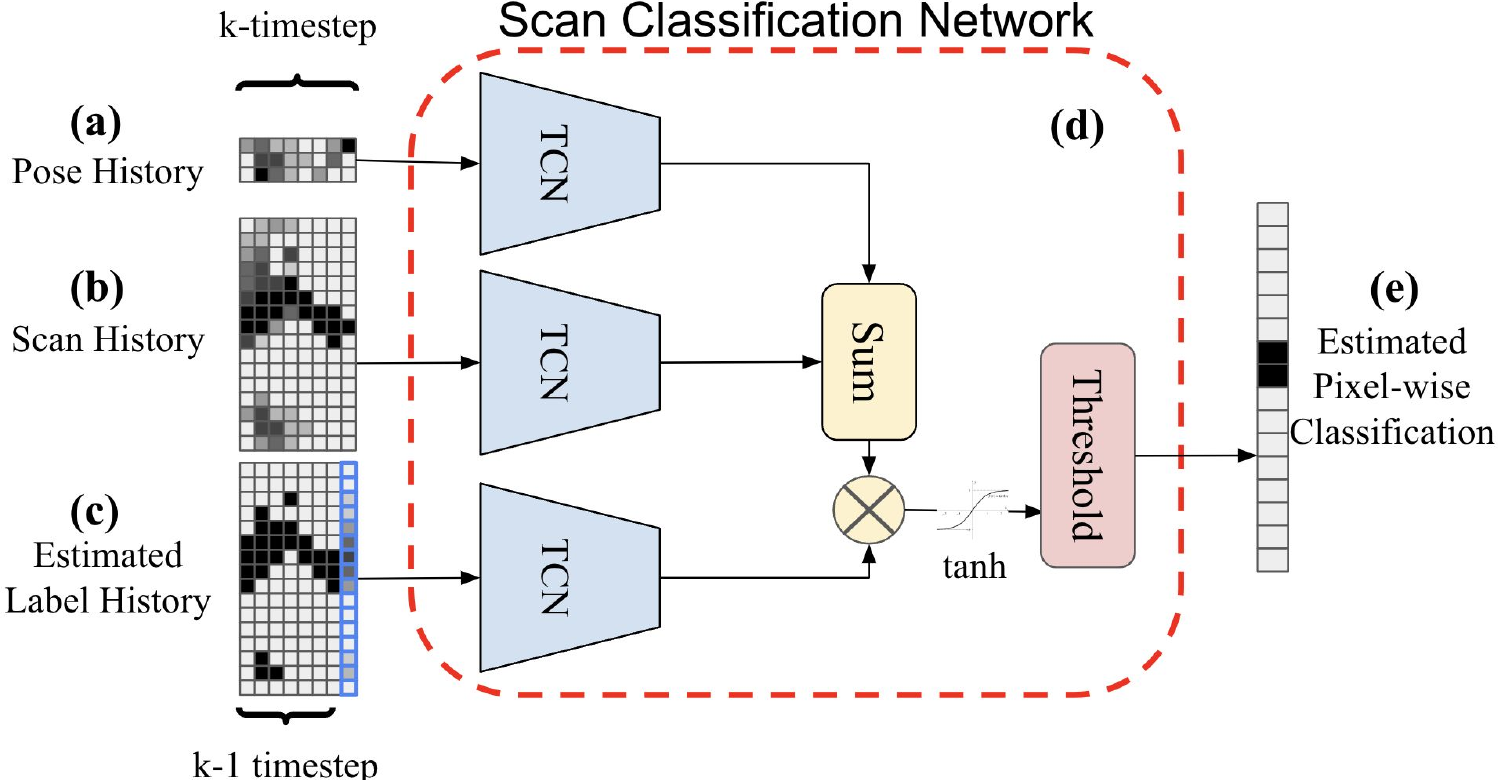}
    \caption{Pixel-wise LiDAR scan classification model architecture used to speed up search by providing information to the planner. $k$ is the length of the history buffer. (a) $[3,k]$ pose history matrix containing $[x,y,\theta]$. (b) $[n_{i},k]$ LiDAR range history matrix. (c) $[n_{i},k-1] \cup [n_{i},1]$ estimated label history matrix concatenated with its pixel-wise exponential weighted average. (d) The model consists of three temporal-convolutional encoders (TCN) (i.e. pose encoder, scan encoder, and label encoder). The encoded poses, scans, and labels are combined to produce a pixel-wise classification of the LiDAR scan. In (e) a threshold (positive/negative) is applied to the raw logits such that each pixel is classified as either a map point or a non-map point.\loosepar{}}
    \label{fig:nn}
\end{figure}

The full model architecture is shown in Fig. \ref{fig:nn} and consists of a temporal-convolutional encoder (TCN) for each input. 
Each TCN contains a single convolutional layer with scan-wise circular padding and a single linear layer. 
The scan and label encoders have kernel size $[k,k]$, and the pose encoder has kernel size $[1,3]$. Hyperbolic tangent activations are used for all layers. 
The output of each TCN is the same size as the LiDAR scan to be classified ($n_i = 897$). 
The output of the pose history TCN is summed with the output of the scan encoder as a pose correction before element-wise multiplication with the label encoding, which may flip the sign of the term. 
The result is normalized by applying a hyperbolic tangent function such that each value is $\{-1,1\}$.
A threshold (positive/negative) is applied to the raw logits such that each pixel is classified as either a map point or a non-map point. The policy is trained with mean squared error loss against ground truth classifications. Training for 20 epochs on a workstation Nvidia 3080 12Gb takes 3 minutes.\loosepar{}

\begin{figure}
    \centering
    \includegraphics[width=\columnwidth]{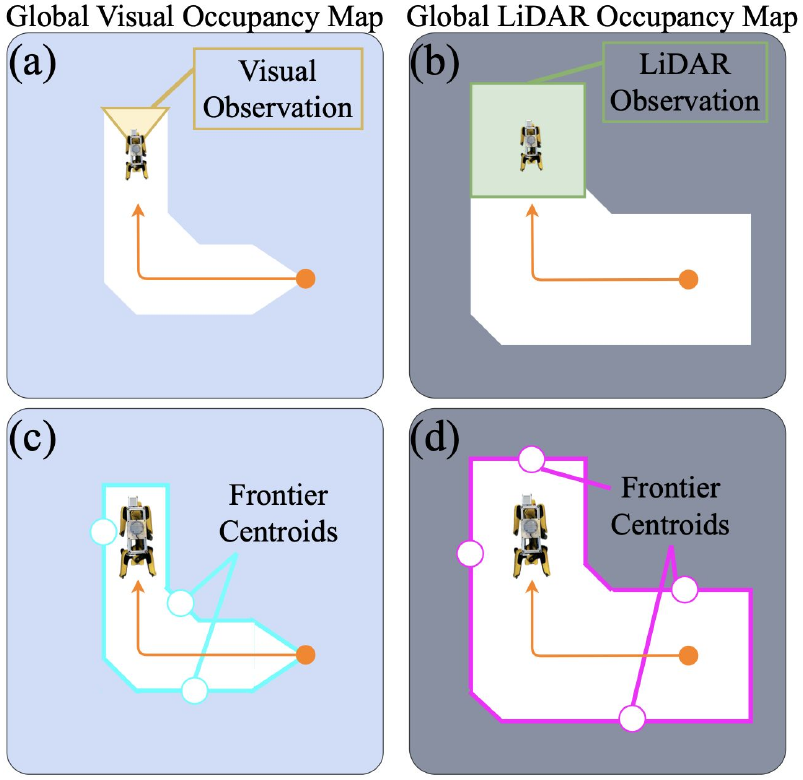}
    \caption{Map updates are performed from observation at each timestep for visual (a) and LiDAR (b) sensors. White cells are known free and blue/grey cells are unknown. Next, frontiers and their centroids are computed for each of the two sensors (c) and (d).\loosepar{}}
    \label{fig:candidate-generation}
\end{figure}

\subsection{Map Updates}

Two global occupancy maps (search maps) are maintained.
One for the LiDAR and the other for the visual sensor (see fig. \ref{fig:candidate-generation}(a-b)).
Global search maps are implemented as occupancy grids.
They are updated at each step from local sensor occupancy grids, which are implemented using the ROS Costmap2D package \cite{costmap}.
All cells in both global maps are initialized to be unknown, $e = -1$ $\forall e \in \mathcal{E}$.
The robot is equipped with a visual sensor with a cone shaped FoV, implemented as a triangular costmap.
The LiDAR sensor has a 360 degree FoV, represented as a square-type costmap.
Observations for both sensors are a set of range estimates to the nearest occupied point $e \in \mathcal{E}^{\text{occ}}$.
Points between the robot and the nearest occupied point along each ray of the LiDAR scan are free points $e \in \mathcal{E}^{\text{free}}$.
Given incoming observations from the robot, each global costmap is updated as free space $e = 0$ or occupied $e = 1$.\loosepar{}

\subsection{Viewpoint Planning}

The planner sends the viewpoint at each planning step that results in the highest likelihood to find the target of interest.
This involves picking viewpoints that balance exploration of unknown space in the environment with exploiting non-map points from LiDAR scans.
This is achieved in several steps: candidate viewpoints are generated, viewpoints are scored, and best sample is fed to the robot.

\subsubsection{Candidate Viewpoint Generation}
\begin{figure}
    \centering
    \includegraphics[width=\columnwidth]{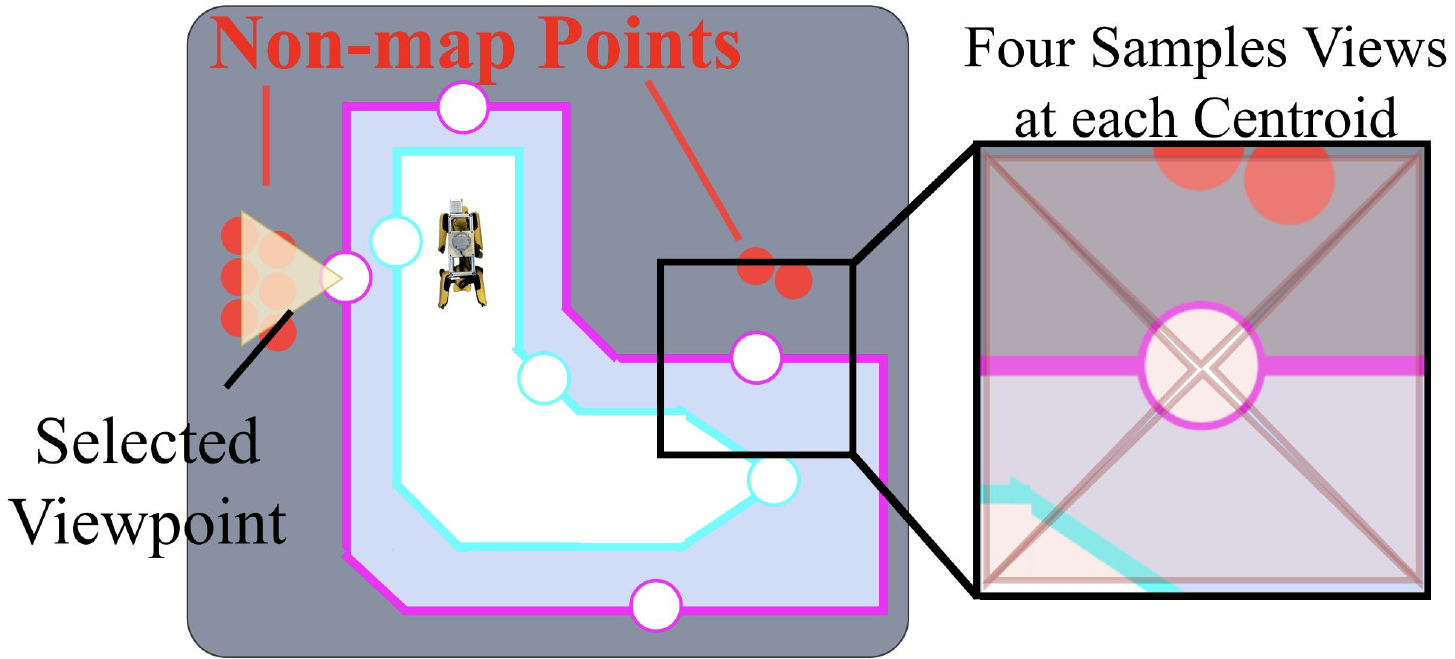}
    \caption{An overview of the viewpoint sampling process. Four viewpoints are considered at each centroid, shown as red triangles. The highest scoring viewpoint is shown as a yellow triangle to be sent to the robot.\loosepar{}}
    \label{fig:viewpoint-selection}
\end{figure}

Frontier points in each of the two global search maps are generated using a process similar to \cite{yamauchi1997frontier}, shown in Fig. \ref{fig:candidate-generation}(c-d).
Any free cell ($e = 0$) adjacent to an unknown cell ($e = -1$) is a frontier edge, which are grouped together into frontiers.
Any group of frontier points beyond a minimum size becomes its own frontier.
Frontier centroids of each frontier in both the visual and LiDAR search maps are found by clustering.
Four candidate viewpoints are evaluated at each frontier centroid to account for the limited FoV of the vision sensor (see fig. \ref{fig:viewpoint-selection}).\loosepar{} 

\subsubsection{Viewpoint Selection}
\label{sec:utility}
Candidate viewpoints are scored using the following utility function.
High value viewpoints are those that are near the robot, allow inspection of many unknown map points and result in inspection of non-map points.
The utility functions is a weighted sum of the following: (1) Penalize distance from robot position to viewpoint, (2) Reward expected number of unknown cells discovered, 
(3) Reward viewpoints with frontiers near the path to the candidate viewpoint, and (4) Reward viewpoints that result in inspection of non-map points.
Scaling parameters on each term are tuned during experiments. 
Weights are set such that the agent is inclined to prioritize non-map points and those viewpoints that result in high number of expected cells discovered. 
The next waypoint is selected by sending the highest utility viewpoint, $x^*_{\text{next}}$, to the robot.\loosepar{}


\section{Results}
\label{sec:results}
We test three hypotheses through our evaluation: (1) The proposed approach to inform an exploring agent with non-map features is effective at finding search targets, (2) it reduces detection time compared against existing algorithms, and (3) it can be deployed successfully on a real robot in unseen indoor environments.\loosepar{}

\begin{table*}
\centering
 \begin{tabular}{||c|c||c|c|c|c|c||} 
 \hline
 \textbf{Map} & \textbf{Target} & \textbf{NBVP} & \textbf{RRT} &  \textbf{MFE}  & \textbf{Ground Truth} & \textbf{Our}  \\ [0.5ex]
\textbf{Name} & \textbf{Difficulty}  & \textbf{\cite{bircher2016receding}} & \textbf{\cite{umari2017autonomous}} & \textbf{\cite{vidal2020multisensor}} &  \textbf{Classification} & \textbf{Method}  \\ [0.5ex]
 \hline
\multirow{2}{*}{Apartment} & Easy & $112 (70\%)$ & $72 (80\%)$ & $\textbf{36 (100\%)}$ & $\textbf{34} \textbf{(100\%)}$ & \color{BrickRed} $\textbf{34} \textbf{(100\%)}$ \\
  & Hard & $177 (30\%) $ & $128 (60\%)$ & $\textbf{106 (100\%)} $  & $\textbf{92} \textbf{(100\%)}$ &  \color{BrickRed} $\textbf{90} \textbf{(100\%)}$ \\
 \hline
  \multirow{2}{*}{Office}& Easy & $70 (80\%)$ & $54 \pm (80\%)$ & $\textbf{34 (100\%)}$ &  $\textbf{26} \textbf{(100\%)}$ & \color{BrickRed} $\textbf{24} \textbf{(100\%)}$\\
  & Hard & $300 (0\%)$ & $\textbf{111 (80\%)}$ & $141 (90\%)$ & $\textbf{99} \textbf{(100\%)}$ & \color{BrickRed} $\textbf{105} \textbf{(100\%)}$ \\
 \hline
 \end{tabular}
  \caption{Average time in seconds (and success rate) for each method to find the object of interest in simulation environments.}
  \label{tab:times}
\end{table*}
\subsection{LiDAR Scan Classification}

We compare the learned classification policy to the ground truth classifier. The output of the policy is a vector of size $n_i$ normalized between -1 and 1 for each scan point. If a value is non-negative, the point belongs to $\mathcal{E}^{\text{map}}$. Otherwise the point belongs to $\mathcal{E}^{\text{non-map}}$. We express performance in terms of per-scan accuracy of the classified LiDAR scans from the policy vs. ground truth from map information. This metric for classified scan $s^{\text{class}}_i$ at time $i$ compared to ground truth $s^{\text{class, true}}_i$ is given by \loosepar{}
\begin{figure}
    \centering
\includegraphics[width=0.9\columnwidth]{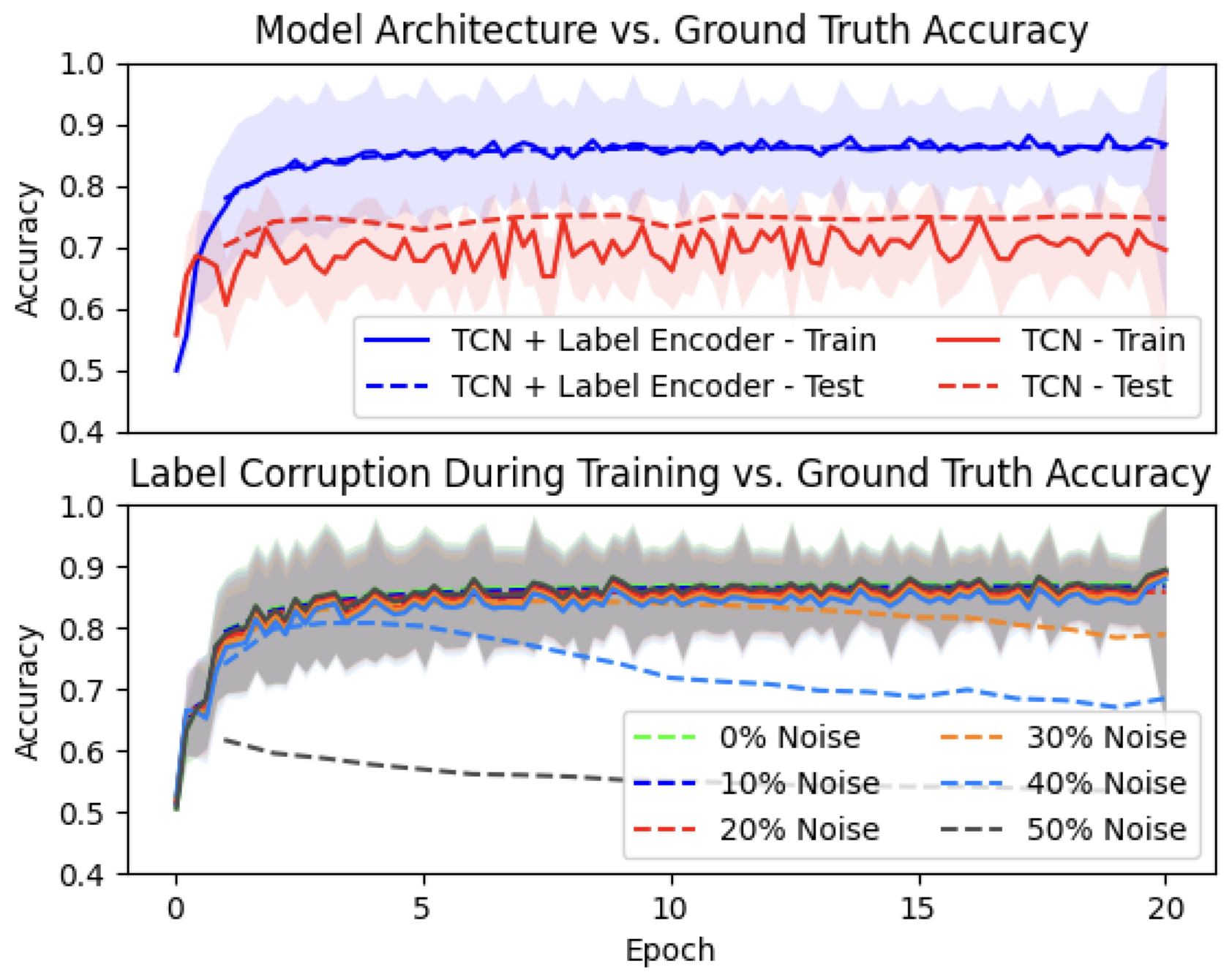}
    \caption{Ablation studies over policy architecture (top) and injected noise during training (bottom). The inclusion of the label history buffer yields 11.63\% higher test accuracy. The policy is robust up to 30\% to bit-flipping errors in the label history buffer. The mean accuracy depicts the 5-step moving average.\loosepar{}}
    \label{fig:ablation}
\end{figure}
\vspace{-0.5em}
\begin{equation}
\text{accuracy} = \frac{1}{{n_i}} \sum^{n_i}_{j=0} \{+1 \text{  if  } p^j_i == p^{j, \text{true}}_i \text{  else  } 0\}
\end{equation}
\vspace{-0.75em}

The final test accuracy of the policy compared to ground truth is $86.19\% \pm 0.03\%$ with no a priori map requirement. We perform two ablation studies, modulating the model architecture and injected noise during training, with results in Fig. \ref{fig:ablation}. The inclusion of the label encoder is responsible for a 11.63\% increase in test accuracy. Because the robot moves relatively slowly compared to the policy update frequency, recently predicted labels strongly bias the current estimate toward the correct classification. We fix the history buffer to 9 time steps, corresponding to 1.8 seconds. The history buffer enables the policy to reason about its environment even though no reference map is provided. During training, uniform random noise is injected into the label history buffer to simulate inference-time auto-regressive accumulated errors. The policy is robust up to 30\% of injected noise before the policy begins to over-fit and performance degrades. We select a policy trained with 10\% injected noise to deploy on the robot. All ablation studies were performed with a fixed seed (0) and initialized with identical weights. We do not tune the random seed.\loosepar{}

\subsection{Simulation and Baseline Comparison}

Simulations are performed in Gazebo with the Turtlebot and ROS navigation.
Baselines are: 1) modified Multisensor Frontier Exploration (MFE) as described in section \ref{sec:related} \cite{vidal2020multisensor}, 2) Next-Best View Planner (NBVP) \cite{bircher2016receding}, 3) Rapidly-Exploring Random Tree (RRT) \cite{umari2017autonomous}, and 4) our method, but informed with ground-truth scan classifications as described in section \ref{sec:ground-truth}. 
\begin{figure}[t]
    \centering
    \includegraphics[width=\columnwidth]{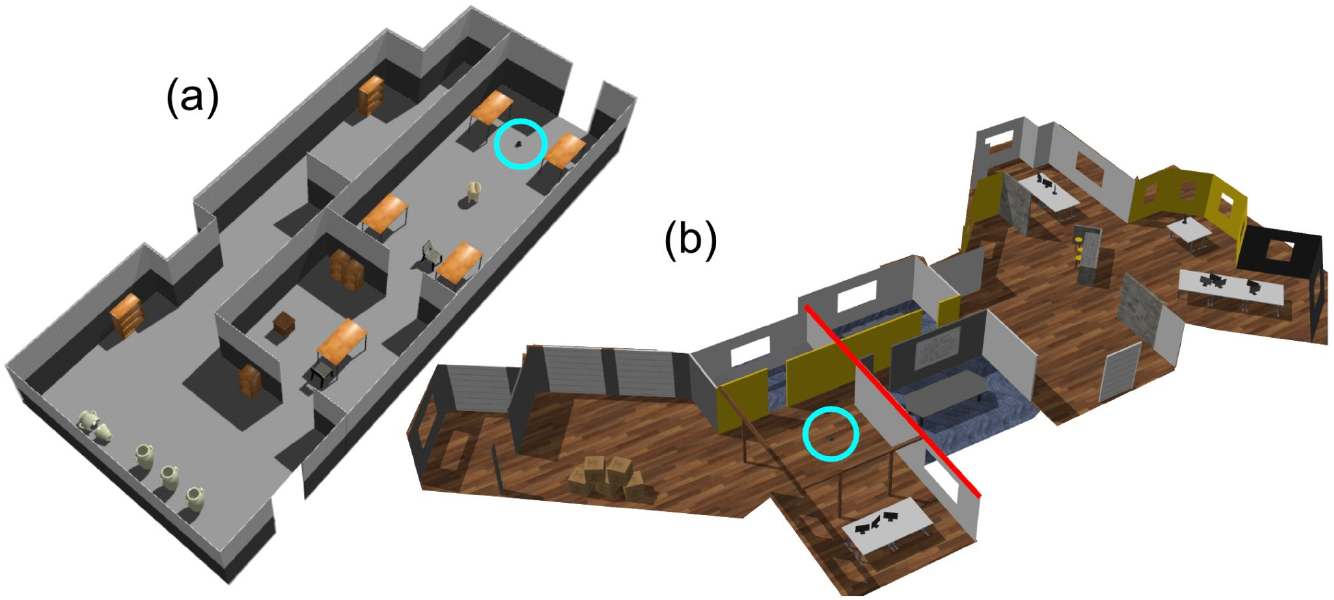}
    \caption{Environments used for baseline comparison. Robot start pose is denoted by a cyan circle. (a) The Apartment (20x30m) simulation environment in the hard configuration. Easy configuration has fewer objects. (b) The Office (25x45m) environment. Easy setup involves the search target to the left of the red line and hard involves the search target on the right.\loosepar{}}
    \label{fig:sim}
\end{figure}
Two Gazebo worlds are tested with easy and hard setups in each (fig. \ref{fig:sim}).
The apartment (20x30m) examines the impact of non-target objects on the planner, the easy setting with fewer objects.
The office (25x45m) tests the method in a large environment under significantly varied target position.
Easy and hard settings are defined by the search object in the same or different half of the environment as the robot starting position.
Both difficulty levels in each map are repeated 10 times.
Average detection times and success rates are reported in Table \ref{tab:times}.
The time limit given to planners is 2 and 3 minutes in the apartment easy and hard settings then 3 and 5 minutes in the office easy and hard settings.
Trials where the planner fails or the time limit is reached are reported as a failure and failed trials are reported as the maximum allotted time.\loosepar{}

Overall, the results support the proposed approach, reporting 100\% success and outperforming baselines across all evaluation settings.
Further, the nearly identical performance between the proposed approach with both predicted and ground truth labels support the efficacy of the perception module.
While MFE performed similarly to the proposed approach in the easy apartment, the lack of the labeled scan data slowed task completion by 18-42\% across the remaining settings.
NBVP and RRT baselines struggled in most settings, with the former averaging about 50\% success.
RRT outperformed other baselines in the hard office environment, however, was 20\% less successful and moderately slower than the proposed approach.
The overall results support our method, suggesting the benefits of focusing on non-map points.
This focus allows the agent to move on when the target is unlikely to be nearby.
These results support hypotheses (1) and (2) that contextual LiDAR information significantly improves performance in the search task.\loosepar{}

\begin{figure}
    \centering    
    \includegraphics[width=0.95\columnwidth]{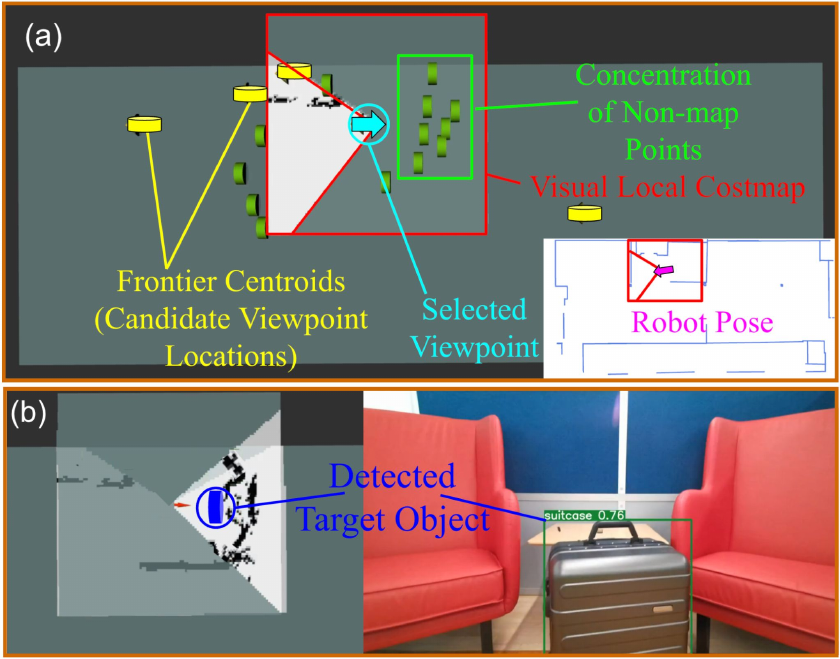}
    \caption{A toy experimental scenario. (a) The first planning step. Candidate viewpoints are sampled at yellow disks. The robot chooses the next viewpoint at the shown cyan arrow in order to visually inspect the concentration of non-map points behind the robot's starting pose. The robot pose on the ground truth map is shown in the lower right (unknown to the robot, not included in the training dataset). (b) The concentration of non-map points is found to be two red chairs and the search object (suitcase). This result reinforces that the scan classifier successfully recognizes non-permanent features in unseen indoor environments.\loosepar{}}
    \label{fig:time-lapse}
\end{figure}

\subsection{Hardware Experiments}

Experiments are in a 20mx30m apartment and a 15mx38m hallway environment. 
These two environments are not included in dataset $\mathcal{D}$ used to train the classification policy. 
A Boston Dynamics Spot is equipped with a Velodyne Puck LiDAR and Azure Kinect RGB-D camera. 
We first evaluate the accuracy of the trained classifier in these out of distribution environments to ground truth information.
This is accomplished by teleoperating the robot while running the ground truth and map-free classifiers in parallel.
The accuracy in the two out of distribution environments are $84.0168\%$ in the apartment over a 105 second deployment and $84.869\%$ in the hallway over a 230 second deployment. 
These suggests the classifier's ability to distinguish permanent features common to many indoor environments. \loosepar{}

During experiments, YOLOv5 \cite{yolov5} is used for detection.
Objects are localized by first converting PointCloud data into the optical frame and then converting to pixel coordinates using camera intrinsics.
Bounding boxes are then used to obtain the depth information from coinciding pixels.
A simple experiment is shown in Fig. \ref{fig:time-lapse} and demonstrates the benefits of non-map information in guiding the vision sensor in the robot's surroundings.
\begin{figure}
    \centering    
    \includegraphics[width=0.95\columnwidth]{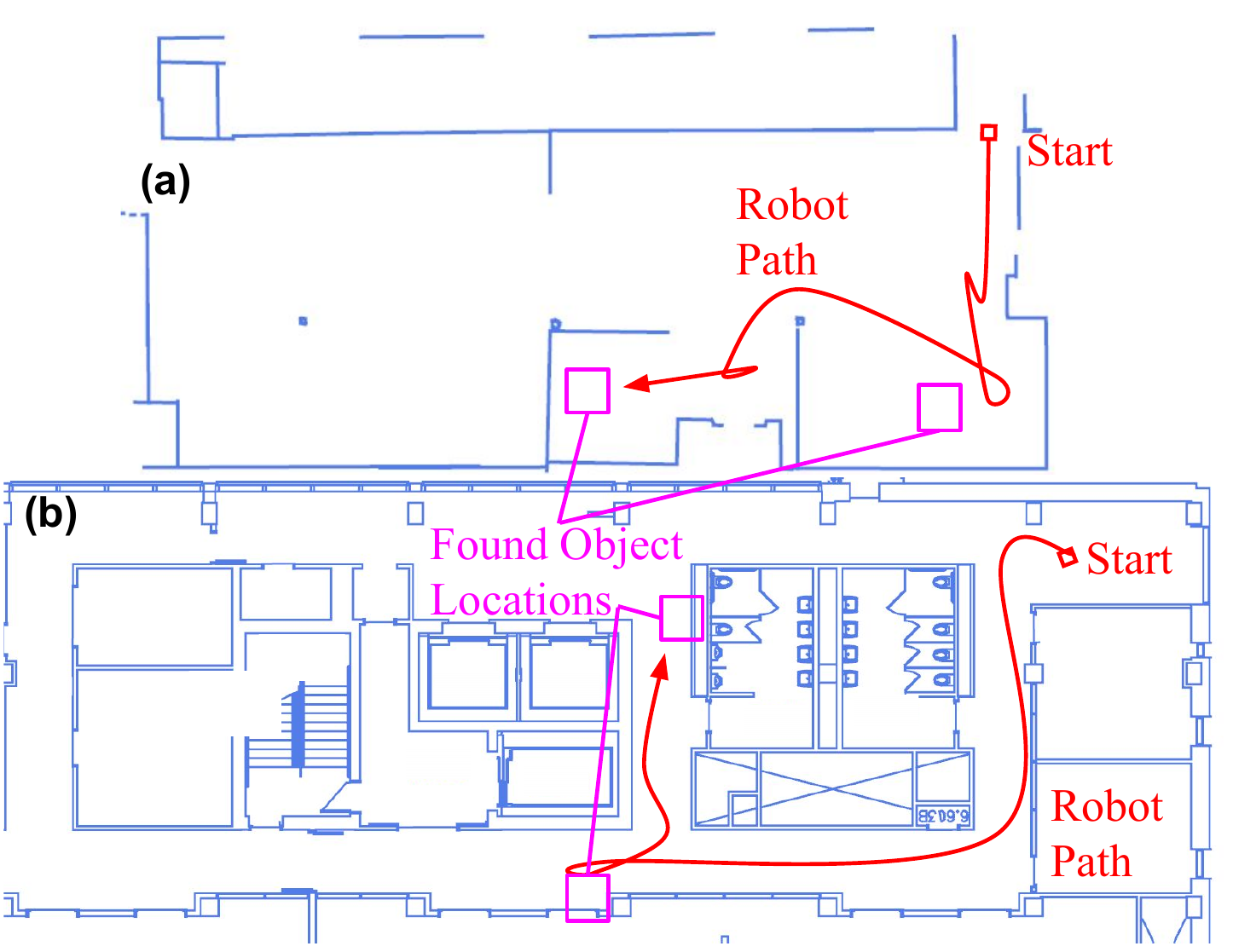}
    \caption{Paths the robot took to find the objects of interest (suitcases) in the video scenarios 2 and 3. (a) Two search objects in the 20x30m apartment environment. (b) Two search objects in the 15x38m hallway environment.\loosepar{}}
    \label{fig:video-paths}
\end{figure}
Instead of selecting based on number of unknown cells, contextual information leads the agent to inspect many non-map features nearby, resulting in detection. 
Videos of experimentation, implementation details, code, training data, and all maps can be found on the project website \url{https://sites.google.com/view/lives-2024/home}.
The video shows two experiments in the apartment and one in the hallway.
Fig. \ref{fig:video-paths} demonstrates the path of the robot in scenarios 2 and 3.
In both apartment scenarios, the robot tends to look towards furniture like couches, shelves and tables.
Similarly, Spot also tends to inspect trash cans and other objects in the hallway.
This behavior demonstrates the impact of non-map points and further supports the efficacy of the segmentation policy.
In the hallway environment the robot is guided by non-map points at several key forks.
After first inspecting the concentration of non-map points situated behind the robot, the decisions the robot makes at the two key forks it intersects lead it to quickly find the two suitcases in the environment.
Overall, the results of hardware experiments on the Spot support hypothesis (3).\loosepar{}


\section{Conclusions}
\label{sec:conclusion}

We introduce a novel visual planner that leverages contextual information found from wide field-of-view LiDAR scans.
The presented map-free LiDAR classifier identifies non-permanent points in the environment for the planner's attention.
The method is shown to outperform baselines by 10-30\% in simulation.
Through ablative studies, we validate our architecture and training approach for map-free scan classification. 
Our experiments confirm the effectiveness of the visual search planner and learning method in real-world scenarios.
Future work plans to leverage the non-map points to generate higher quality maps and expand to multi-robot applications.\loosepar{}

\section{Acknowledgments}

This research was supported in part by NSF Award \#2219236 (GCR: Community Embedded Robotics: Understanding Sociotechnical Interactions with Long-term Autonomous Deployments) and Living and Working with Robots, a core research project of Good Systems, a UT Grand Challenge. Any opinions, findings, and conclusions or recommendations expressed in this material are those of the author and do not necessarily reflect the views of the National Science Foundation.



\printbibliography

@inproceedings{saroya2020online,
  title={Online exploration of tunnel networks leveraging topological CNN-based world predictions},
  author={Saroya, Manish and Best, Graeme and Hollinger, Geoffrey A},
  booktitle={2020 IEEE/RSJ International Conference on Intelligent Robots and Systems (IROS)},
  pages={6038--6045},
  year={2020},
  organization={IEEE}
}

@inproceedings{ramakrishnan2020occupancy,
  title={Occupancy anticipation for efficient exploration and navigation},
  author={Ramakrishnan, Santhosh K and Al-Halah, Ziad and Grauman, Kristen},
  booktitle={Computer Vision--ECCV 2020: 16th European Conference, Glasgow, UK, August 23--28, 2020, Proceedings, Part V 16},
  pages={400--418},
  year={2020},
  organization={Springer}
}

@article{caley2019deep,
  title={Deep learning of structured environments for robot search},
  author={Caley, Jeffrey A and Lawrance, Nicholas RJ and Hollinger, Geoffrey A},
  journal={Autonomous Robots},
  volume={43},
  pages={1695--1714},
  year={2019},
  publisher={Springer}
}

@article{biswas2017episodic,
  title={Episodic non-markov localization},
  author={Biswas, Joydeep and Veloso, Manuela M},
  journal={Robotics and Autonomous Systems},
  volume={87},
  pages={162--176},
  year={2017},
  publisher={Elsevier}
}

@inproceedings{biswas2012depth,
  title={Depth camera based indoor mobile robot localization and navigation},
  author={Biswas, Joydeep and Veloso, Manuela},
  booktitle={2012 IEEE International Conference on Robotics and Automation},
  pages={1697--1702},
  year={2012},
  organization={IEEE}
}

@inproceedings{best2022resilient,
  title={Resilient multi-sensor exploration of multifarious environments with a team of aerial robots},
  author={Best, Graeme and Garg, Rohit and Keller, John and Hollinger, Geoffrey A and Scherer, Sebastian},
  booktitle={Robotics: Science and Systems (RSS)},
  year={2022}
}

@article{vidal2020multisensor,
  title={Multisensor online 3D view planning for autonomous underwater exploration},
  author={Vidal, Eduard and Palomeras, Narc{\'\i}s and Isteni{\v{c}}, Klemen and Gracias, Nuno and Carreras, Marc},
  journal={Journal of Field Robotics},
  volume={37},
  number={6},
  pages={1123--1147},
  year={2020},
  publisher={Wiley Online Library}
}

@article{kaelbling1998planning,
  title={Planning and acting in partially observable stochastic domains},
  author={Kaelbling, Leslie Pack and Littman, Michael L and Cassandra, Anthony R},
  journal={Artificial intelligence},
  volume={101},
  number={1-2},
  pages={99--134},
  year={1998},
  publisher={Elsevier}
}

@inproceedings{nguyen2024grey,
  title={Grey Wolf Optimization-Based Path Planning for Unmanned Aerial Vehicles in Bridge Inspection},
  author={Nguyen, LV and Le, TH and Ha, Quang Phuc},
  booktitle={2024 IEEE/SICE International Symposium on System Integration (SII)},
  pages={810--815},
  year={2024},
  organization={IEEE}
}

@inproceedings{yamauchi1997frontier,
  title={A frontier-based approach for autonomous exploration},
  author={Yamauchi, Brian},
  booktitle={Proceedings 1997 IEEE International Symposium on Computational Intelligence in Robotics and Automation CIRA'97.'Towards New Computational Principles for Robotics and Automation'},
  pages={146--151},
  year={1997},
  organization={IEEE}
}

@inproceedings{charrow2015information,
  title={Information-theoretic mapping using cauchy-schwarz quadratic mutual information},
  author={Charrow, Benjamin and Liu, Sikang and Kumar, Vijay and Michael, Nathan},
  booktitle={2015 IEEE International Conference on Robotics and Automation (ICRA)},
  pages={4791--4798},
  year={2015},
  organization={IEEE}
}

@inproceedings{bircher2016receding,
  title={Receding horizon "next-best-view" planner for 3D exploration},
  author={Bircher, Andreas and Kamel, Mina and Alexis, Kostas and Oleynikova, Helen and Siegwart, Roland},
  booktitle={2016 IEEE International Conference on Robotics and Automation (ICRA)},
  pages={1462--1468},
  year={2016},
  organization={IEEE}
}

@inproceedings{umari2017autonomous,
  title={Autonomous robotic exploration based on multiple rapidly-exploring randomized trees},
  author={Umari, Hassan and Mukhopadhyay, Shayok},
  booktitle={2017 IEEE/RSJ International Conference on Intelligent Robots and Systems (IROS)},
  pages={1396--1402},
  year={2017},
  organization={IEEE}
}

@inproceedings{dai2020fast,
  title={Fast frontier-based information-driven autonomous exploration with an mav},
  author={Dai, Anna and Papatheodorou, Sotiris and Funk, Nils and Tzoumanikas, Dimos and Leutenegger, Stefan},
  booktitle={2020 IEEE international conference on robotics and automation (ICRA)},
  pages={9570--9576},
  year={2020},
  organization={IEEE}
}

@article{zheng2023system,
  title={A System for Generalized 3D Multi-Object Search},
  author={Zheng, Kaiyu and Paul, Anirudha and Tellex, Stefanie},
  journal={arXiv preprint arXiv:2303.03178},
  year={2023}
}

@misc{yolov5,
author = {Jocher, Glenn},
doi = {10.5281/zenodo.3908559},
license = {AGPL-3.0},
month = may,
title = {{YOLOv5 by Ultralytics}},
url = {https://github.com/ultralytics/yolov5},
version = {7.0},
year = {2020},
note={}
}

@article{liang2022reconnaissance,
  title={A reconnaissance penetration game with territorial-constrained defender},
  author={Liang, Li and Deng, Fang and Wang, Jianan and Lu, Maobin and Chen, Jie},
  journal={IEEE Transactions on Automatic Control},
  volume={67},
  number={11},
  pages={6295--6302},
  year={2022},
  publisher={IEEE}
}

@article{christiansen2017designing,
  title={Designing and testing a UAV mapping system for agricultural field surveying},
  author={Christiansen, Martin Peter and Laursen, Morten Stigaard and J{\o}rgensen, Rasmus Nyholm and Skovsen, S{\o}ren and Gislum, Ren{\'e}},
  journal={Sensors},
  volume={17},
  number={12},
  pages={2703},
  year={2017},
  publisher={Multidisciplinary Digital Publishing Institute}
}

@article{chen2023hybrid,
  title={A Hybrid Planning Method for 3D Autonomous Exploration in Unknown Environments With a UAV},
  author={Chen, Xuning and Zheng, Jianying and Hu, Qinglei},
  journal={IEEE Transactions on Automation Science and Engineering},
  year={2023},
  publisher={IEEE}
}

@article{bi2023cure,
  title={CURE: A Hierarchical Framework for Multi-Robot Autonomous Exploration Inspired by Centroids of Unknown Regions},
  author={Bi, Qingchen and Zhang, Xuebo and Wen, Jian and Pan, Zhangchao and Zhang, Shiyong and Wang, Runhua and Yuan, Jing},
  journal={IEEE Transactions on Automation Science and Engineering},
  volume={99},
  pages={1--14},
  year={2023},
  publisher={IEEE}
}

@article{wang2019autonomous,
  title={Autonomous robotic exploration by incremental road map construction},
  author={Wang, Chaoqun and Chi, Wenzheng and Sun, Yuxiang and Meng, Max Q-H},
  journal={IEEE Transactions on Automation Science and Engineering},
  volume={16},
  number={4},
  pages={1720--1731},
  year={2019},
  publisher={IEEE}
}

@misc{costmap,
author={Marder-Eppstein, Eitan 
 and Lu, David V. and Hershberger, Dave},
url={http://wiki.ros.org/costmap_2d},
year={2024},
note={}
}

@misc{enml, author={Biswas, Joydeep}, url={https://github.com/ut-amrl/enml},
journal={ros.org},
note={}}

@misc{rrt, author={Umari, Hassan}, url={https://github.com/hasauino/rrt_exploration}, journal={ros.org}, year={2024},
note={}}

@article{chen2022automatic,
  title={Automatic labeling to generate training data for online LiDAR-based moving object segmentation},
  author={Chen, Xieyuanli and Mersch, Benedikt and Nunes, Lucas and Marcuzzi, Rodrigo and Vizzo, Ignacio and Behley, Jens and Stachniss, Cyrill},
  journal={IEEE Robotics and Automation Letters},
  volume={7},
  number={3},
  pages={6107--6114},
  year={2022},
  publisher={IEEE}
}

@article{jhaldiyal2023semantic,
  title={Semantic segmentation of 3D LiDAR data using deep learning: a review of projection-based methods},
  author={Jhaldiyal, Alok and Chaudhary, Navendu},
  journal={Applied Intelligence},
  volume={53},
  number={6},
  pages={6844--6855},
  year={2023},
  publisher={Springer}
}

@article{qi2017pointnet,
  title={Pointnet++: Deep hierarchical feature learning on point sets in a metric space},
  author={Qi, Charles Ruizhongtai and Yi, Li and Su, Hao and Guibas, Leonidas J},
  journal={Advances in neural information processing systems},
  volume={30},
  year={2017}
}

@article{zamorski2020adversarial,
  title={Adversarial autoencoders for compact representations of 3D point clouds},
  author={Zamorski, Maciej and Zieba, Maciej and Klukowski, Piotr and Nowak, Rafal and Kurach, Karol and Stokowiec, Wojciech and Trzci{\'n}ski, Tomasz},
  journal={Computer Vision and Image Understanding},
  volume={193},
  pages={102921},
  year={2020},
  publisher={Elsevier}
}

@article{mersch2022receding,
  title={Receding moving object segmentation in 3d lidar data using sparse 4d convolutions},
  author={Mersch, Benedikt and Chen, Xieyuanli and Vizzo, Ignacio and Nunes, Lucas and Behley, Jens and Stachniss, Cyrill},
  journal={IEEE Robotics and Automation Letters},
  volume={7},
  number={3},
  pages={7503--7510},
  year={2022},
  publisher={IEEE}
}
\balance 

\end{document}